\documentclass[runningheads]{llncs}
\usepackage{graphicx}

\usepackage{mathtools}
\usepackage{floatrow}  
\usepackage{threeparttable}

\newfloatcommand{capbtabbox}{table}[][\FBwidth]
\DeclareFloatSeparators{mysep}{\hskip1em}

\begin{document}
%===========================================================

\title{SCPNet: Spatial-Channel Parallelism Network for Joint Holistic and Partial Person Re-Identification} % Replace your paper's title here
\titlerunning{SCPNet} % Replace an abstracted version of your paper's title here

%===========================================================

\author{Xing Fan\inst{1,2} \and
Hao Luo\inst{1,2} \and
Xuan Zhang\inst{2} \and
Lingxiao He\inst{3} \and
Chi Zhang\inst{2} \and
Wei Jiang\inst{1}}
%
%Please include author names in full in the paper, 
%If any authors have names that can be parsed into FirstName LastName in multiple ways, please include the correct parsing, in a comment to the volume editors:
%\index{Lastnames, Firstnames}

\authorrunning{X. Fan et al.} % A shorter version of authors' name

%===========================================================

\institute{
    Zhejiang University, Hangzhou, China\\
    \email{\{xfanplus,haoluocsc,jiangwei\_zju\}@zju.edu.cn}\and
    Megvii Inc. (Face++), Beijing, China\\
    \email{\{zhangxuan,zhangchi\}@megvii.com} \and
    Inst. of Automation, Chinese Academy of Sciences, Beijing, China\\
    \email{lingxiao.he@nlpr.ia.ac.cn}
}

\maketitle

%===========================================================
\begin{abstract}
Holistic person re-identification (ReID) has received extensive study in the past few years and achieves impressive progress. However, persons are often occluded by obstacles or other persons in practical scenarios, which makes partial person re-identification non-trivial. In this paper, we propose a spatial-channel parallelism network (SCPNet) in which each channel in the ReID feature pays attention to a given spatial part of the body. The spatial-channel corresponding relationship supervises the network to learn discriminative feature for both holistic and partial person re-identification. The single model trained on four holistic ReID datasets achieves competitive accuracy on these four datasets, as well as outperforms the state-of-the-art methods on two partial ReID datasets without training.

\end{abstract}

%===========================================================
\section{Introduction}

\begin{figure}
	\centering
	\includegraphics[width=.8\linewidth]{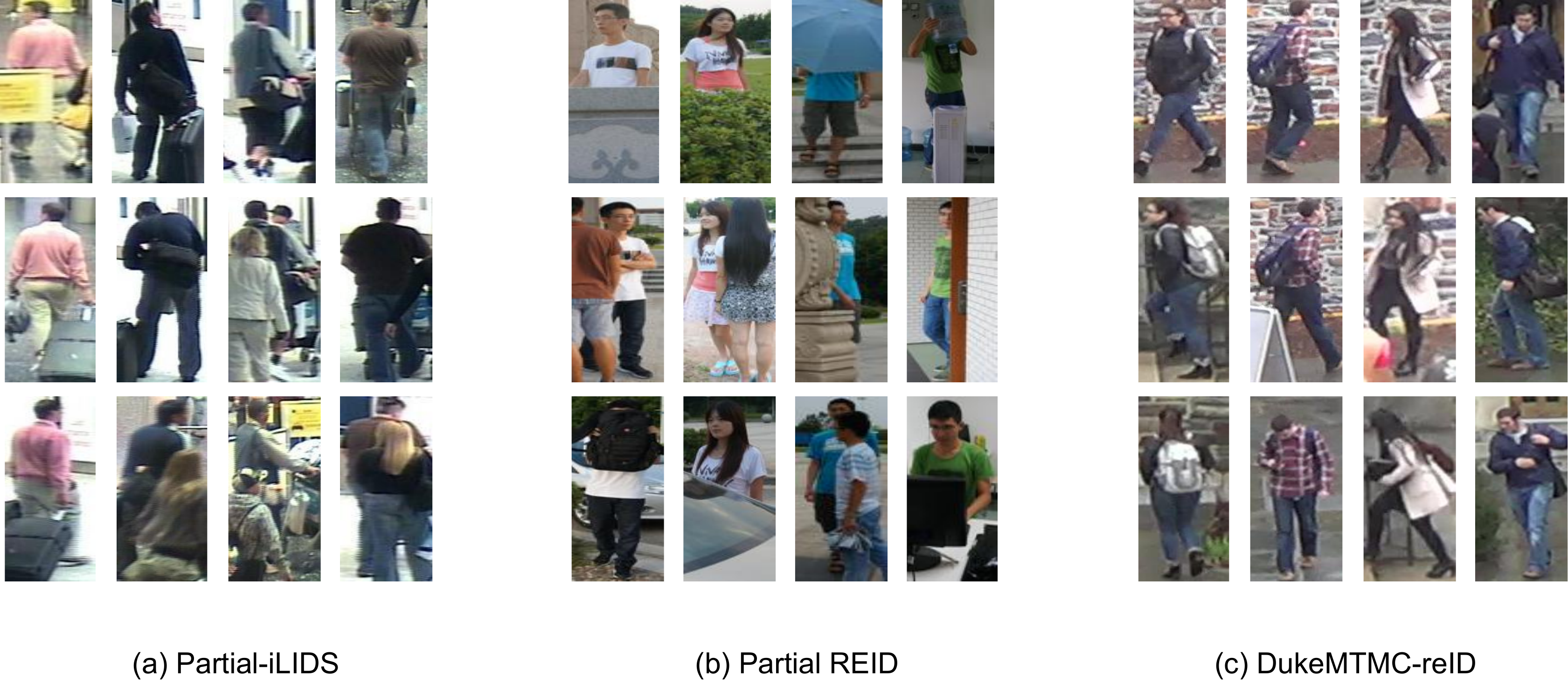}
	\caption{Samples of holistic person and partial person images.}
	\label{fig:sample}
\end{figure}

Person re-identification (ReID) is a popular research problem in computer vision.
However, existing works always focus on the holistic person images, which cover a full glance of one person.
There are some holistic person images shown in Fig. \ref{fig:sample}(c).
In the realistic scenario, the person may be occluded by some moving or static obstacles (e.g. cars, walls, other persons), as shown as in Fig. \ref{fig:sample}(a) and Fig. \ref{fig:sample}(b).
Therefore, partial person ReID is an important issue for real-world ReID applications and has gradually attracted researchers' attention.
The occlusions will change the feature of person appearance, which presents a big challenge to identify a person across views.
There are few studies focusing on partial person ReID, while most current holistic person based approaches can not well solve the partial person images.
From this perspective, studying partial person ReID, especially jointing holistic person ReID, is necessary and crucial both for academic research and practical ReID applications.

Most of current ReID studies use representation learning \cite{lin2017improving,zheng2016person} or metric learning \cite{cheng2016person,chen2017beyond,hermans2017defense,liu2017end,xiao2017margin} to learn a global feature, which is sensitive to the occlusions changing the person appearance.
Some works, which use part or pose guided alignment \cite{sun2017beyond,yao2017deep,Zhao_2017_ICCV,wei2017glad,zhao2017spindle,zheng2017pose}to learn local features, can boost the performance of holistic person ReID, but may fail to get good alignment information when the person images are non-holistic.
To overcome the above problems, several partial ReID methods have been proposed in recent years.
Sliding Window Matching (SWM) \cite{Zheng_2015_ICCV} introduces an alternative solution for partial ReID by setting up a sliding window of the same size as the probe image and using it to search for the most similar region within each gallery person.
However, for SWM, the size of probe person is smaller than the size of the gallery person.
Some part-to-part matching (PPM) based methods divide the image into many parts of the same size and use local part-level features to solve partial ReID.
The searching process of SWM and PPM based methods is time-consuming.
Hence \cite{he2018deep} proposed Deep Spatial feature Reconstruction (DSR), which exploits the reconstruction error of two images of different sizes to leverage Fully Convolution Network (FCN).
Replacing part-based searching with direct reconstruction, DSR accelerates the process of ReID.
However, DSR needs partial ReID data to train a good partial ReID model.

In this paper, we propose an end-to-end model named spatial-channel parallelism network (SCPNet), which is only trained on holistic ReID datasets but performs well on both holistic and partial person ReID datasets.
In the proposed framework, SCPNet includes a global branch and a local branch.
As same as most of traditional methods, the global branch uses the global average pooling (GAP) to extract the global feature.
To acquire the local feature, we divide feature maps into several equal parts from top to down and then apply the horizontal pooling to get local features of each part.
Inspired by the motivation that we expect the global feature can store certain local information, we design a loss to leverage global branch through the local feature, which is the output of the local branch.
Due to this design, SCPNet can learn how to do partial ReID and keep working in holistic person ReID at the same time without partial ReID training data.
In previous research works, combing the global feature with the local feature is a common solution to boost the performance.
However, the dimension of the feature vector determines the speed of retrieval process.
SCPNet can only use the output features of the global branch to halve the dimension and speed up retrieval, because the features contain the local information.
In the following, we overview the main contents of our method and summarize the contributions:

\begin{itemize}
	\item {We }propose a novel approach named Spatial-Channel Parallelism Network (SCPNet) for both holistic and partial person ReID, which effectively use the local features to leverage the global features in the training phase.
	\item{Besides, }our SCPNet can perform better on partial ReID with only being trained on holistic ReID datasets, which let it more suitable for the practical ReID scene.
	\item {Experimental} results demonstrate that the proposed model achieves state-of-the-art results on four holistic ReID datasets. And our unsupervised cross-domain results of partial ReID beat the supervised state-of-the-art results by a large margin on Partial REID \cite{Zheng_2015_ICCV} and Partial-iLIDS \cite{zheng2011person} datasets. 
\end{itemize}

%===========================================================
\section{Related Works}

Since the proposed approach joints holistic and partial person ReID using deep convolutional networks, we briefly introduce some related deep learning based works in this section.

\textbf{Representation learning.} 
Deep representation learning which regards ReID as a classification task such as verification or identification problem, is a commonly supervised learning method in person ReID \cite{lin2017improving,zheng2016person,SphereReID}.
In \cite{zheng2016person}, Zheng \emph{et al.} make the comparison between verification baseline and identification baseline:
1) For the former, a pair of input person images is judged whether they belong to the same person by a deep model.
2) For the latter, the method treats each identity as a category, and then minimizes the softmax loss.
In some improved works \cite{lin2017improving}, person attributes loss is combined with verification or identification loss to learn a better feature description.

\textbf{Metric learning.} 
In deep metric learning, the deep model can directly learn the similarity of two images according to the L2 distance of their feature vectors in the embedding space.
The typical metric loss includes contrastive loss, triplet loss and quadruplet loss in terms of the training pairs.
Usually, two images of the same person are defined as a positive pair, whereas two images of different persons are a negative pair.
Contrastive loss minimizes the features distance of the positive pairs while maximizing the features distance of the negative pairs.
However, triplet loss \cite{liu2017end} is motivated by the margin enforced between positive and negative pairs.
A triplet only has two identities, and quadruplet adds an extra identity to get a new negative pair.
In addition, selecting suitable samples for the training model through hard mining has been shown to be effective \cite{hermans2017defense,xiao2017margin}.
Combining softmax loss with metric learning loss to speed up the convergence is also a popular method \cite{geng2016deep}. 

\textbf{Part-based methods.}
Part-based methods are commonly used to extract local features in holistic person ReID.
This kind of methods has very great inspiration for occlusion problem in partial person ReID.
In details, human pose estimation and landmark detection having achieved impressive progress, several recent works in ReID employ these tools to acquire aligned subregion of person images \cite{wei2017glad,zhao2017spindle,zheng2017pose,AlignedReID2017}.
Then, using a deep model extract the spatial local features of each part is an effective way to boost the performance of global features.
Another common solution is to divide images into several parts without an alignment \cite{sun2017beyond,yao2017deep,Zhao_2017_ICCV} and concatenate the features of each part.
However, when occlusion happened, the above methods are usually inefficient because of the incomplete pose points or disturbed local features.
In addition, they usually concatenate the global and local features to avoid losing the global information, which means they need extra extracted global features.
The SCPNet is inspired by using the local features to leverage the global features, and is suitable for both holistic and partial person ReID.

\textbf{Partial Person ReID.}
Solving the occlusion problem in person ReID is important for the ReID applications.
However, there have been few methods which consider how to learn a better feature to match an arbitrary patch of a person image.
In some other tasks, some works \cite{donahue2014decaf,girshick2014rich} easily wrap an arbitrary patch of an image to a fixed-size image, and then extract the fixed-length feature vector for matching.
In \cite{Zheng_2015_ICCV}, Zheng \emph{et at.} improved it into a global-to-local matching model named Sliding Window Matching (SWM) that can capture the spatial layout information of local patches, and also introduced a  local patch-level matching model called Ambiguity-sensitive Matching Classifier (AMC) that was based on a sparse representation classification formulation with explicit patch ambiguity modeling.
However, the computation cost of AMC-SWM is expensive because it extracts feature using much time without sharing computation.
He \emph{et at.} proposed a method \cite{he2018deep} that leveraged Fully convolutional Network (FCN) to generated certain-sized spatial feature maps such that pixel-level features are consistent.
Then, it used Deep Spatial feature Reconstruction (DSR) to match a pair of person images of different sizes.
DSR need to solve an optimization problem to calculate the similarity of two feature maps of different size, which increases the time consumption in application.
To this end, we propose a model called SCPNet which can output a fixed-length feature vector to solve both holistic and partial ReID.

%===========================================================
\section{Our Proposed Approach}

\begin{figure}[htb]
	\centering
	\includegraphics[width=.7\linewidth]{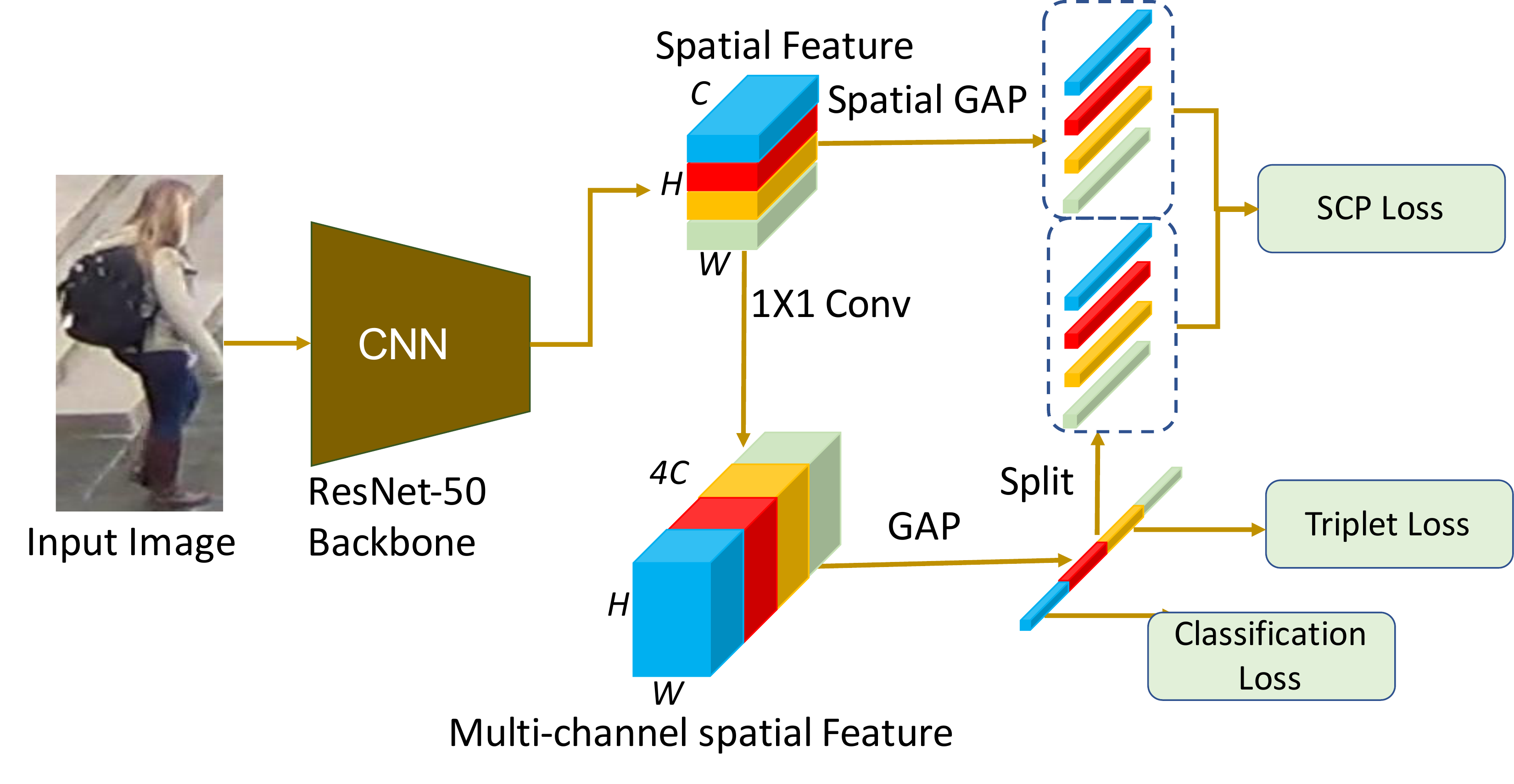}
	\caption{Structure of our SCPNet model. We feed input images forward into the ResNet-50\cite{he2016deep} backbone network and extract output $C \times H \times W$ feature map s of last convolutional layer. Then we divide a feature map into four spatial horizontal stripes and apply global average pooling (GAP) for each stripe producing local features. We also apply an $1\times1$ convolutional layer on the feature map to produce multi-channel spatial feature representation, and then we use GAP to obtain the compact feature. The compact feature is split into four feature vector parts by the channel. Finally, three losses are computed.}
	\label{fig:network}
\end{figure}

In this section, we present our spatial-channel parallelism network (SCPNet) as in Fig. \ref{fig:network}.

In SCPNet, a single global feature is generated for the input image, and the L2 distance of such features are used as similarity in the inference stage.
However, the global feature is learned jointly with local features in the learning stage.

For each image, we use a CNN, such as ResNet50 \cite{he2016deep}, to extract a feature map as the original appearance representation, which is the output of the last convolution layer ($C \times H \times W$, where $C$ is the channel number and $H \times W$ is the spatial size. On the other hand, an $1 \times 1$ convolutional layer is applied to the feature map to extend the channel number from $C$ to $RC$, where $R$ is the number of body region ($R=4$ in Fig. \ref{fig:network}), then a global pooling is applied to extract the global feature (a $RC$-d vector). On the other hand, the feature map is uniformly divided into $R$ parts in vertical direction, and a global pooling is applied for each of them to extract $R$ local features ($C$-d vectors). 

Intuitively, the global feature represents the whole person, while $R$ local features represent different body regions. However, both of these features have drawbacks if trained independently. For the global feature, it often focuses on certain part and ignores other local details. For the local features, as demonstrated in \cite{luo_understanding_2016}, an effective receptive field only occupies a fraction of the full theoretical receptive field and a lot of local information is still preserved after many convolution layers. Since their effective receptive fields are small, they lack enough context information to well represent the corresponding body region. Moreover, because of misalignment, such as pose variation and inaccurate detection boxes, a local feature may not correspond to the body regions accurately. 

To alleviate the drawbacks, we propose a spatial-channel parallelism method. After obtaining the global feature and local features, each local feature is mapped to a part of the global feature (with the same color as in Fig. \ref{fig:network}). In more details, the $r$-th local feature, which is a $C$-dimensional vector, should be closer to the continuously $C$ channels from the $rC$-th to the $(r+1)C$-th in the global feature.
We use the local features to let each part of the global feature focus on certain body regions by introducing a spatial-channel parallelism loss as follows:

\begin{align}
L_{SCP} = \sum_{r=1}^{R}\| f_{s,r}-f_{c,r}\|_2^2
\label{eq:L_featue}
\end{align}
where $f_{s,r}$ is the $r$-th local feature, and $f_{c,r}$ is the $r$-th part of the global feature (the $rC$-th to the $(r+1)C$-th channel). 

\begin{figure}[htb]
	\centering
	\includegraphics[width=.5\linewidth]{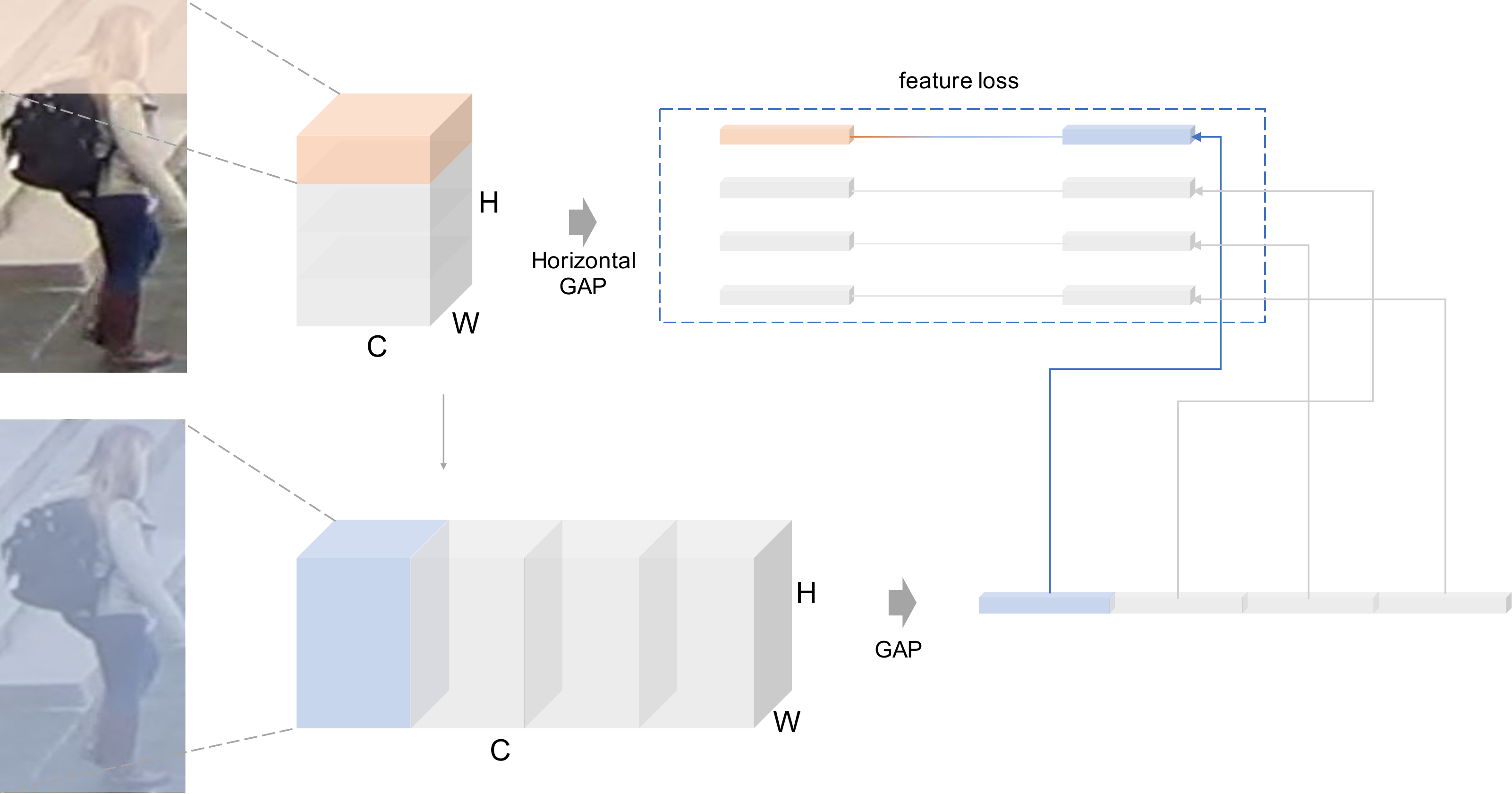}
	\caption{The example of spatial-channel parallelism. The feature map of the global branch is produced from the entire original feature map of the local branch. 
		Each channel part of the global branch has global receptive filed (blue region) and contains global information of the whole input image, while the first part of the local branch maps to the orange region in the image.}
	\label{fig:comparison}
\end{figure}

The two types of features should be close, because it is actually a re-learning process which enforces every channel part to learn the corresponding local spatial feature, but from the entire feature map, see Fig. \ref{fig:parts}(b). The misalignment can be handled because the corresponding part is extracted automatically from whole input even when the relevant part appears in other position of the input image. Thus the resulting re-learned feature is a more robust representation.

In this way, with the supervision of local features, each part of the global feature is forced to learn local representation of the corresponding part of the input. Hence each region of the body is focused by certain channels in the global feature, which makes the global feature hold more local details and more discriminative. Moreover, different from local features, each part of the global feature has a global receptive field, which leads to better representation of the corresponding local region. It also has no rigid spatial segmentation, thus the global feature is more robust to pose variation and inaccurate bounding boxes.

In the learning stage, the loss is comprised of three losses: the metric learning loss $L_{metric}$, the classification loss $L_{class}$ and the spatial-channel parallelism loss $L_{SCP}$. For the metric learning loss, the similarity between two images is defined by the L2 distance of their global features, and the TriHard loss proposed by \cite{hermans2017defense}  is chosen, where for each sample, the most dissimilar one with the same identity and the most similar one with a different identity is chosen to obtain a triplet. For the classification loss, a softmax is applied for each image with the identity. The spatial-channel parallelism loss is given in Eq. (\ref{eq:L_featue}), which requires local features to compute.
And the final loss is the sum of above three loss as following:
\begin{align}
L = L_{class}+L_{metric}+\lambda L_{SCP}
\label{eq:L_all}
\end{align}
where $\lambda$ is the weight of spatial-channel parallelism loss.

In the inference stage, only the global feature is needed. For a holistic person, its ReID feature is represented by the whole global feature, while for a partial person, its ReID feature is represented by part of the global feature. Hence, the similarity of two person images is computed as the L2 distance of the shared part of their global features. For both holistic and partial ReID, we use the same feature vector and let SCPNet to learn to determine which part of feature should be extracted
automatically.

\section{Experiments}
\label{section:experiments}

\subsection{Datasets}
We train a single model end-to-end on four challenging holistic ReID datasets, including Market-1501\cite{zheng2015scalable}, DukeMTMC-reID\cite{ristani2016MTMC}, CUHK03\cite{Li2014DeepReID} amd CUHK-SYSU\cite{xiao2016end}, and then test on these four datasets. Furthermore, we also directly test the trained model on two partial ReID datasets including Partial REID\cite{Zheng_2015_ICCV} and Partial-iLIDS\cite{zheng2011person} without training.

\textbf{Market-1501} consists of 32,668 bounding box images from 1,501 person identities captured by six cameras in front of a supermarket.The provided pedestrian bounding boxes are detected by Deformable Part Model (DPM)\cite{felzenszwalb2010object}. The training set consists of 12,936 images from 751 identities and testing set contains the other 19,732 images from 750 identities.

\textbf{DukeMTMC-reID} consists of 36,411 images from 1,812 person identities captured by 8 high-resolution cameras. There are 1,404 identities appear in more than two cameras and the other 408 identities are regarded as distractors. Training set consists of 702 identities and testing set contains the rest 702 identities.

\textbf{CUHK03} consists of 14,096 images from 1,467  person identities captured by six cameras in the CUHK campus. Both manually labeled pedestrian bounding boxes and automatically detected bounding boxes are provided. In this paper, we use the manually labeled version.

\textbf{CUHK-SYSU} contains 18,184 full images and 99,809 bounding boxes from 8,432 person identities are provided. The dataset is divided into training set containing 11,206 images of 5,532 identities and testing set containing the other 6,978 images of 2,900 identities.

\textbf{Partial REID} contains 600 images of 60 people, with 5 full-body images and 5 partial images per person. The images were collected at an university campus with different viewpoints, background and different types of severe occlusions. 

\textbf{Partial-iLIDS} is a simulated partial person datasets based on i-LIDS\cite{zheng2011person}. There are 119 people with total 476 person images captured by multiple non-overlapping cameras. Some images occlusion, sometimes rather severe, caused by people and luggage.

\subsection{Implementation Detials}
We implement our propose SCPNet model using the PyTorch framework. The backbone network is the ResNet-50\cite{he2016deep} model pre-trained on ImageNet. In training phase, the input image is resized to $288\times144$ then randomly cropped to $256\times128$ with random horizontal flip. The mini-batch size is set to 64, in which each identity has 4 images. Before feeding the input image into the network, we subtract the mean value and then divide it by the standard deviation as same as the normalization procedure of the ResNet-50\cite{he2016deep} trained on ImageNet.

We use the Adam optimizer with the default hyper-parameters ($\epsilon=10^{-8}$, $\beta_1=0.9$, $\beta_2=0.999$) to minimize the network loss. The initial learning rate is $10^{-3}$, and we lower the learning rate twice at epoch 80 and 180 to $10^{-4}$ and $10^{-5}$ respectively. The total training takes 300 epochs. Weight decay is set to $10^{-5}$ and never changes. We also add a dropout layer for classification and the dropout ratio is set to 0.75 in all our experiments.

We will release the code and trained weights of our SCPNet model after publication and more details can be found in the code.

\subsection{Results}

In this section, we focus on five aspects: 1) The Ablation study on spatial-channel parallelism.
2) The influence of number of parts.
3) The influence of SCP loss weight.
4) State-of-the-arts results of holistic ReID.
5) State-of-the-arts results of partial ReID.

For partial ReID datasets, similar to Market-1501\cite{zheng2015scalable}, we provide top-$k$ accuracy by finding the most similar correct match in the top $k$ candidates.

\subsubsection{Ablation Study.}

To evaluate the benefits of the proposed SCPNet, we compare it with a baseline network.
The baseline network is the same as SCPNet, but without the SCP loss in the learning stage.
For the baseline network, there are three features can be extracted as the ReID feature in the experiments:
1) the global feature, which is the output of the global branch, 2) the local feature, which is the concatenation of all features in the local branch, 3) and the concat feature, which is the concatenation of the previous two features.
For SCPNet, we set $R =4$ and $\lambda=10$ in the experiments. 

As Table. \ref{table:different features} shown, we report the rank-1 accuracy of four experiments on totally six ReID datasets mentioned. 
For the baseline network, we report the results when the global/local/concatenate feature is used as the ReID features.
It is easy to see that the concatenate feature is superior to both the global and the local feature in the baseline network.
However, the SCPNet always outperforms the baseline network with the concat feature on the ReID datasets.
It is shown that let the global feature and the local feature learn each other by spatial-channel parallelism is better than simply concatenating them. Moreover, the channel number used as ReID feature in SCPNet is only half of that in the baseline using the concat feature.

\begin{table}[htb]\scriptsize
	\begin{center}
		\caption{Influence of different features. When SCP loss is not used, we apply softmax and triplet loss on spatial feature branch, channel feature branch or both branches respectively. When SCP loss is used, we use structure shown in Fig.~\ref{fig:network} and extract different output as final representation.}
		\label{table:different features}
		\renewcommand{\arraystretch}{1.2}
		\begin{tabular}{c|c|c|c|c|c|c|c}
			\hline
			\multicolumn{2}{c|}{Settings} & \multicolumn{6}{c}{Results (r=1), $R=4, \lambda = 10$} \\
			\hline
			Model & Branch & Market-1501 & DukeMTMC-reID & CUHK03 & CUHK-SYSU& Partial ReID & Partial-iLIDs \\
			\hline
			&Local  & 90.7 & 77.5 & 90.3 & 94.5 & 61.0 & 83.2 \\
			Baseline		&Global & 92.0 & 80.2 & 89.6 & 93.1 & 60.0 & 78.2 \\
			&Concate & 92.8 & 81.7 & 92.7 & 93.6 & 61.0 & \textbf{84.9} \\
			\hline
			\multicolumn{2}{c|}{SCPNet}& \textbf{94.1} & \textbf{84.8} & \textbf{93.3} & \textbf{94.6} & \textbf{68.3} & \textbf{84.9}  \\
			\hline
		\end{tabular}
	\end{center}
\end{table}

On the partial ReID task, the local branch performs better than the global branch on Partial REID and Partial-iLIDS datasets, which indicates that local features are more important for partial ReID.
And our SCPNet exceeds the global branch of \emph{Baseline} by more than $6.0\%$ rank-1 accuracy ($1.3\% \sim 4.6\%$ rank-1 accuracy for holistic ReID).
It can be learned that our proposed SCPNet improves much more performance in partial ReID than holistic ReID.
In addition, the SCPNet is not inferior to the concatenate branch, which includes local and global features.
In other words, the local information is transferred from the local branch to the global branch.
The results of ablation study show the effectiveness of our proposed approach.

\subsubsection{Influence of number of parts.}

Because the size of the output feature map of the SCPNet is $8 \times 4$ in our experiments, we can divide the feature map into $1,2,4,8$ parts respectively. As shown as Table. \ref{table:different parts}, when $R =1$, two branches are both global branches and the SCPNet achieves worst performance on all six datasets.
When we set $R>1$, the SCPNets perform better than the SCPNet ($R=1$) on all datasets, which shows that the local features are beneficial to ReID.

\begin{table}[htb]
	\begin{center}
		\caption{Influence of different number of parts (rank-1 accuracies, $\lambda = 1$).}
		\label{table:different parts}
		\renewcommand{\arraystretch}{1.2}
		\begin{tabular}{c|c|c|c|c|c|c}
			\hline
			$R$  & Market-1501  & DukeMTMC-reID & CUHK03 & CUHK-SYSU & Partial ReID &Partial-iLIDS\\
			\hline
			1  & 92.0 & 80.8 & 89.9 & 93.8 & 61.3 & 82.4 \\
			2 & 93.3 & \textbf{82.9} & \textbf{90.9} & \textbf{94.2} & 65.3 & \textbf{85.7}\\
			4 & \textbf{93.4} & 82.0 & 90.3 & 93.9 & \textbf{67.0} & 84.9 \\
			8 & 93.2 & 82.6 & 90.9 & 94.1 & 65.0 & 84.0 \\
			\hline
		\end{tabular}
	\end{center}
\end{table}

\begin{figure}[ht]
	\centering
	\includegraphics[width=0.6\linewidth]{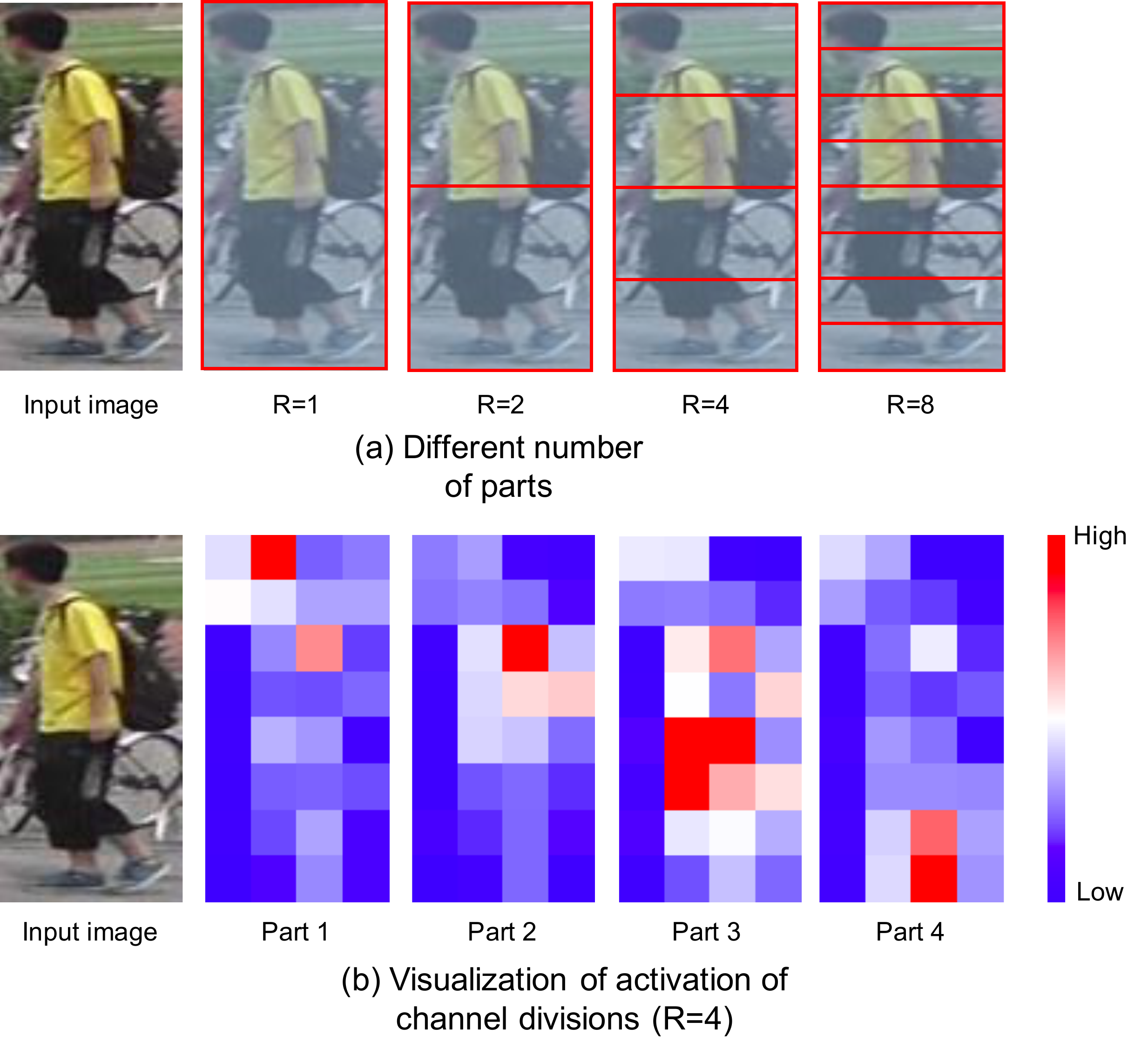}
	\caption{Strip horizontal division. SCPNet with few strips cannot use local information well. SCPNet with more strips cannot preserve the semantic information well (e.g., a "half head" when $R=8$). (b) Spatial-channel parallelism. For every position, we calculate the maximum of all channels belong to this part of the global feature. It shows that each part focuses mainly on the corresponding spatial part.}
	\label{fig:parts}
\end{figure}

SCPNet with few strips will become a common global feature model, which cannot fully utilize the spatial-channel parallelism with more local feature. SCPNet with more strips cannot preserve semantic information of person components, which influences the performance. e.g., Fig. \ref{fig:parts}(a)(R=8), more strips would divide the head and bag into two parts. The experimental result suggests that SCPNet with 4 strips can make full use of local information, and preserve more semantic information. Fig. \ref{fig:parts}(a) illustrates the different strip horizontal division.

We choose R=4 to do the subsequent experiments in this paper.

\subsubsection{Influence of SCP loss weight.}

We fix $R=4$ and use different weights ($\lambda$) of SCP loss to do the experiments.
As shown as Table. \ref{table:feature weight}, when $\lambda = 0$, the model (\emph{Baseline} mentioned previously) gets the lowest rank-1 accuracy on all datasets.
In overall, the performance of the SCPNet is gradually improving when $\lambda$ increases from $0.5$ to $10$, which demonstrates the effectiveness of the spatial-channel parallelism again.
If we continue to increase $\lambda$, the SCPNet will not improve performance again on five datasets except Partial-iLIDS.
When $R=4$ and $\lambda$ is around $10$, the proposed SCPNet achieves the best performance.

\begin{table}[tb]
	\begin{center}
		\caption{Influence of SCP loss weight (rank-1 accuracies, $R=4$).}
		\label{table:feature weight}
		\renewcommand{\arraystretch}{1.2}
		\begin{tabular}{c|c|c|c|c|c|c}
			\hline
			$\lambda$	& Market-1501 & DukeMTMC-reID & CUHK03 & CUHK-SYSU & Partial ReID &Partial-iLIDS\\
			\hline
			0     & 92.0 & 80.2 & 89.6 & 93.1 & 60.0 & 78.2 \\
			0.5 & 93.2 & 81.7 & 90.0 & 94.0 & 63.3 & 83.2 \\ 
			1     & 93.4 & 82.0 & 90.3 & 93.9 & 67.0 & 84.9 \\
			2    & 93.6 & 82.6 & 91.2 & 94.2 & 65.3 & 84.9 \\
			4    & 93.8 & 83.6 & 92.2 & 94.4 & 66.0 & 84.0 \\
			6    & \textbf{94.1} & 83.9 & 92.3 & 94.3 & 64.7 & 84.9 \\
			8    & 93.8 & 83.8 & 92.4 & 94.3 & 68.0 & 84.0 \\
			10  & \textbf{94.1} & \textbf{84.8} & \textbf{93.3} & \textbf{94.6} & \textbf{68.3} & 84.9 \\
			12  & 93.9 & 83.3 & 92.2 & 94.6 & 64.0 & \textbf{86.6} \\
			16  & 93.5 & 83.6 & 92.5 & 94.6 & 67.3 & 82.3 \\
			\hline
		\end{tabular}
	\end{center}
\end{table}

\begin{figure}[t]
	\centering
	\includegraphics[width=.62\linewidth]{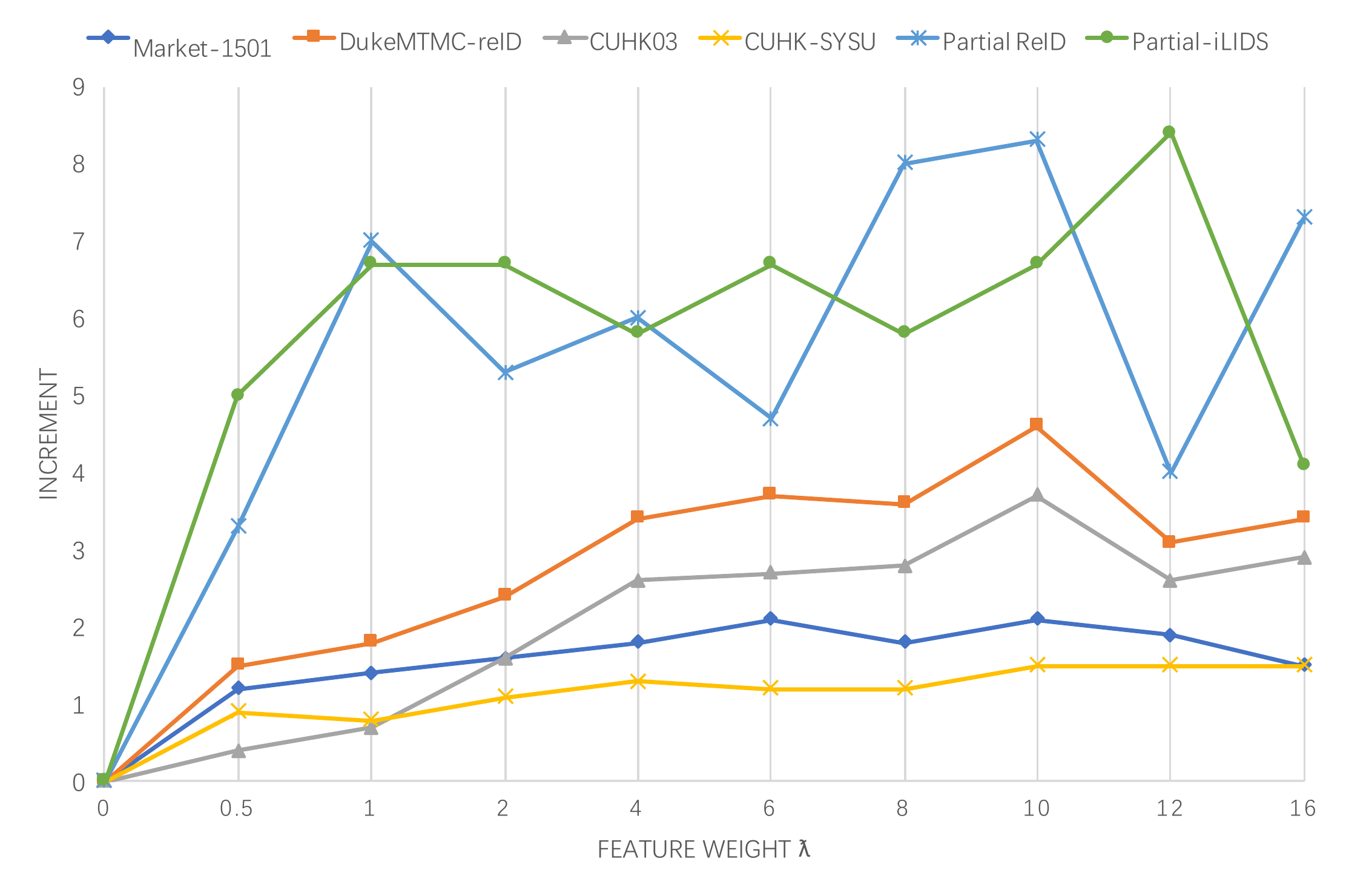}
	\caption{The increments of rank-1 accuracy on six datasets. We use the results of SCPNet ($\lambda = 0$) as the baseline and compute the increment of SCPNet results with different $\lambda$.}
	\label{fig:lambda}
\end{figure}

\subsubsection{Comparison with single-dataset setting.}
Unlike many existing methods, we train a single model on all four datasets. For a fair comparison, without any extra modification or hyperparameter searching, we also train our SCPNet on each dataset alone using exactly the same setting.
As shown in Table~\ref{table:each_dataset}, SCPNet-s is inferior to SCPNet-a, but it still outperforms the baseline by a large margin.

With adjustments especially for single-dataset setting, our SCPNet-s can acquire even better performance. For example, SCPNet-s can achieve $91.7\%$ by changing mini-batch size from 64 to 48 on Market-1501.

\begin{table}[thb]
	\begin{center}
		\caption{Comparison of single model and dataset-specific model. We just report rank-1 accuracies. A method with ``-s" means that the model is trained with corresponding single dataset, while ``-a" means the model is trained with all four datasets. Baseline is the global branch without SCP loss.}
		\label{table:each_dataset}
		\renewcommand{\arraystretch}{1.2}
		\begin{tabular}{c|c|c|c|c}
			\hline
			Method  & Market-1501  & DukeMTMC-reID & CUHK03 & CUHK-SYSU \\
			\hline
			%Local-s  & - & - & - & - \\
			Baseline-s & 88.1 & 77.6 & 89.5 & 90.7 \\
			%Concate-s & - & - & - & - \\
			\hline
			SCPNet-s & 91.2 & 80.3 & 90.7 & 91.9 \\
			SCPNet-a & 94.1 & 84.8 & 93.3 & 94.6 \\
			\hline
		\end{tabular}
	\end{center}
\end{table}

\begin{figure}[bht]
    \CenterFloatBoxes
    \begin{floatrow}
        \ffigbox[\FBwidth]{
            \includegraphics[width=.9\linewidth]{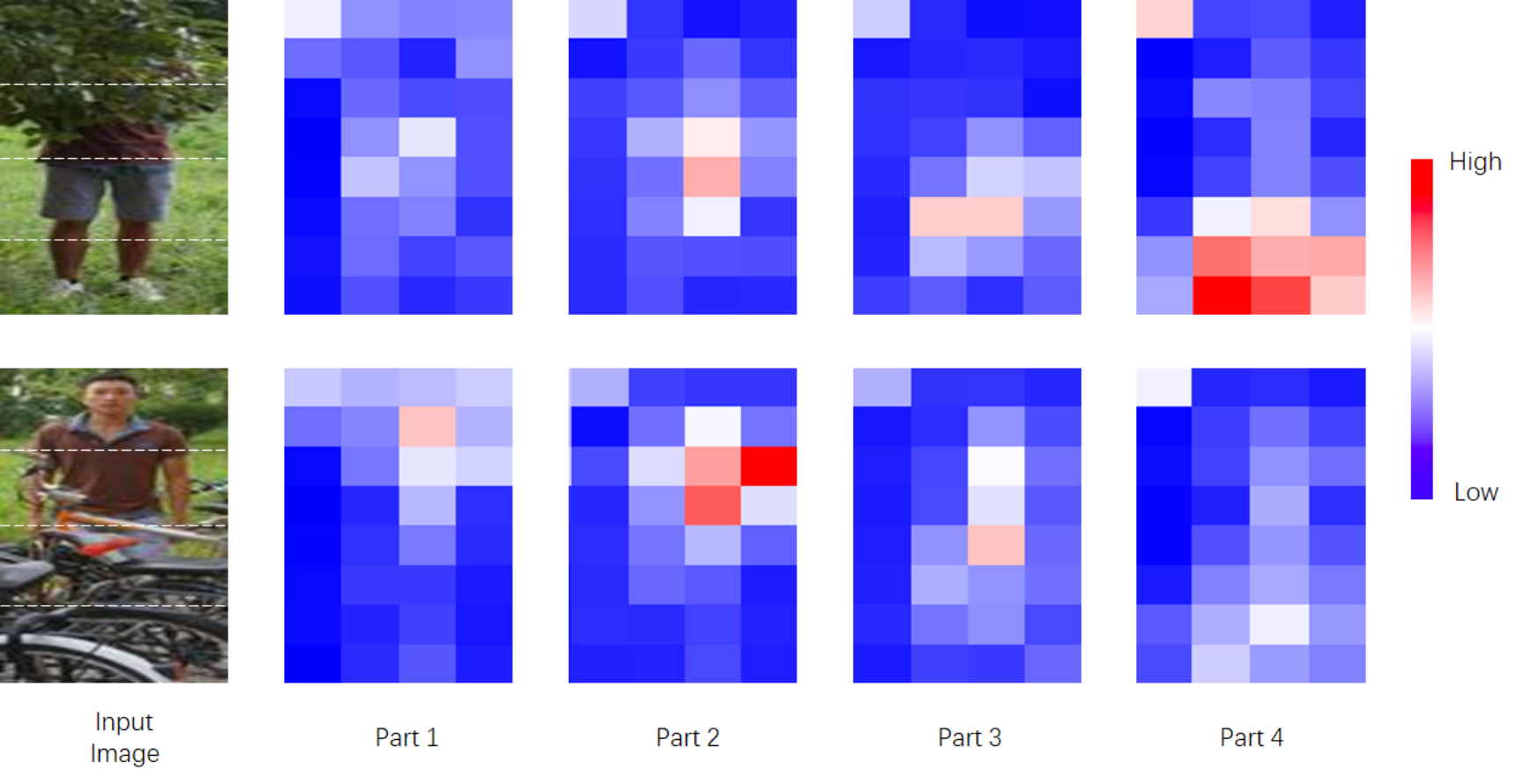}
        }{%
            \caption{Occlusion visualization of activation of channel divisions (R=4)}
            \label{fig:vis_occlusion}
        }
        \killfloatstyle
        \ttabbox[\Xhsize]{
            %\scriptsize
            \begin{tabular}{c|c|c}
                \hline
                Method  & Market-1501  & DukeMTMC-reID \\
                \hline
                Baseline-s & 56.0 & 26.6 \\
                SCPNet-s & 68.1 & 41.6 \\
                \hline
            \end{tabular}
        }{%
            \caption{Trained on CUHK-SYSU then tested on Market-1501 and DukeMTMC-reID directly.}
            \label{tab:trans}
        }
    \end{floatrow}
\end{figure}

The proposed SCP loss can enforce the model to learn more discriminative features. When some region is occluded, feature in other regions is still available. As shown in Table~\ref{tab:trans}, when trained on CUHK-SYSU then tested on Market-1501 and DukeMTMC-reID, SCPNet outperforms baseline significantly (only 1.2\% gap on CHUK-SYSU, but 12.1\% and 15.0\% gap on Market-1501 and DukeMTMC-reID respectively), which implies it has learned to extract richer feature and is more robust to occlusions and appearance variation. Furthermore, SCPNet can focus on human body rather than occlusions. As shown in Fig.~\ref{fig:vis_occlusion}, when the upper or bottom part is occluded, the activation of distractor parts is suppressed, which makes SCPNet suitable for partial ReID task. Misalignment is also handled because we extract each part of feature from whole regions rather than rigid local pooling regions. SCP loss cannot well deal with the occlusion in the vertical direction. We admit that this is a disadvantage, and may introduce vertical stripes into SCP loss in the future.

\subsubsection{Results on Holistic ReID.}

\begin{table} [htb]
	\floatsetup{floatrowsep=mysep}
	\begin{floatrow}  
		\ttabbox
		{
			\begin{tabular}{c|ccc}
				\hline
				method & mAP & r = 1 &r = 5\\
				\hline
				\hline
				LDNS \cite{zhang2016learning}		&35.7	&61.0	&-\\
				Gated S-CNN\cite{varior2016gated}			&39.6	&65.9	&-	\\
				IDNet+VeriNet \cite{chen2017person}			&45.5	&71.8	&-	\\
				Re-ranking \cite{zhong2017re}			&63.6	&77.1	&-\\
				PIE \cite{zheng2017pose}			&56.0	&79.3	&94.4	\\
				XQDA+SSM \cite{bai2017scalable}			&68.8	&82.2	&-\\
				TriHard \cite{hermans2017defense}			&69.1	&84.9	&-	\\
				Spindle\cite{zhao2017spindle}			&-		&76.9	&91.5\\
				CamStyle\cite{zhong2017camera}		&68.7	&88.7	& -\\
				GLAD\cite{wei2017glad}         			&73.9   	&89.9 	&-\\
				HA-CNN\cite{li2018harmonious}		&75.5	&91.2 	&-\\
				\hline
				DSR\cite{he2018deep}				&64.3	&83.6	&-\\
				\hline
				SCPNet-s & 75.2 & 91.2 & 97.0 \\
				SCPNet-a & \textbf{81.8} & \textbf{94.1} & \textbf{97.7} \\
				\hline
			\end{tabular}
		}
		{
			\caption{Experimental results on the Market-1501 with single query}
			\label{table:Market-1501}
		}
		
		\ttabbox
		{
			\begin{tabular}{c|ccc}
				\hline
				Methods  							& r = 1  	& r = 5	&r = 10		\\
				\hline
				\hline
				LOMO+XQDA \cite{liao2015person} 			& 44.6	& -		&-\\
				LDNS \cite{zhang2016learning}		& 62.6	&90.0 	&94.8 \\
				Gated S-CNN\cite{varior2016gated} 			& 61.8	& -		&-\\
				LSTM Simaese \cite{varior2016a} 					& 57.3  	& 80.1	&88.3\\
				Re-ranking \cite{zhong2017re}			& 64.0	& -		&-\\
				PIE \cite{zheng2017pose}			&67.1	&92.2	&96.6	\\
				TriHard \cite{hermans2017defense} 			& 75.5	&95.2	&99.2\\
				OIM \cite{xiao2017joint}				& 77.5	& -		&-\\
				Deep \cite{geng2016deep}			& 84.1	& -		&-\\
				SOMAnet \cite{barbosa2017looking}		& 72.4	&95.2	&95.8\\
				IDNet+VeriNet \cite{zheng2016discriminatively}		& 83.4	& 97.1	&98.7\\
				GLAD\cite{wei2017glad}        			&85.0  	&97.9       &99.1\\
				Spindle\cite{zhao2017spindle}			& 88.5	&97.8	&98.6\\
				\hline
				SCPNet-s			& 90.7	& 98.0 & 99.0	\\
				SCPNet-a			& \textbf{93.3}	& \textbf{98.7}& \textbf{99.2}	\\
				\hline
			\end{tabular}
		}
		{
			\caption{Experimental results on the CUHK03 with detected dataset}
			\label{table:CUHK03}
		}
	\end{floatrow}  
\end{table}

\begin{table} [htb]
	\floatsetup{floatrowsep=mysep}
	\begin{floatrow}  
		\ttabbox
		{
			\begin{tabular}{c|ccc}
				\hline
				method & mAP &r = 1 & r = 5\\
				\hline
				\hline
				BoW+KISSME\cite{zheng2015scalable} 		& 12.2	&25.1  	&-\\
				IDNet \cite{zheng2016person}				&45.0	&65.2 	& -\\
				TriHard \cite{hermans2017defense}				&53.5	&72.4	&-	\\
				SVDNet\cite{sun2017svdnet}				&56.8	&76.7 	&86.4\\
				CamStyle\cite{zhong2017camera}			&57.6	&78.3	&-\\
				HA-CNN\cite{li2018harmonious}			&63.8	&80.5	&-\\
				GP-reID\cite{almazan2018re}				&\textbf{72.8}	&\textbf{85.2} 	&\textbf{93.9}\\
				\hline
				SCPNet-s & 62.6 & 80.3 & 89.6 \\
				SCPNet-a & 68.5 & 84.8 & 91.9 \\
				\hline
			\end{tabular}
		}
		{
			\caption{Experimental results on the DukeMTMC-reID}
			\label{table:DukeMTMC-reID}
		}
		
		\ttabbox
		{
			\begin{tabular}{c|ccc}
				\hline
				Methods  						& mAP	& r = 1	&r = 5		\\
				\hline
				\hline
				VGG16+RSS\cite{xiao2016end}			& 55.7	&62.7	&-		\\
				DLDP \cite{schumann2016deep}	&74.0	&76.7 	&-\\
				NPSM \cite{Liu_2017_ICCV} 		&77.9 	&81.2	&-\\
				\hline
				SCPNet-s & 90.0 & 91.9 & 96.9 \\
				SCPNet-a & \textbf{93.1} & \textbf{94.6} & \textbf{98.0} \\
				\hline
			\end{tabular}
		}
		{
			\caption{Experimental results on the CUHK-SYSU}
			\label{table:CUHK-SYSU}
		}
	\end{floatrow}  
\end{table} 

We compare our SCPNet to existing state-of-the-art methods on four holistic datasets, including Market-1501, CUHK03, DukeMTMC-ReID, and CUHK-SYSU.
Because these four datasets are not overlapping, we train only one single model with all training samples.
For Market-1501, CUHK03, DukeMTMC-ReID, and CUHK-SYSU, we mainly report the mAP and CMC accuracy as same as the standard evaluation.
For CUHK03, because we train one single model for all benchmarks, it is slightly different from the standard procedure in \cite{Li2014DeepReID}, which splits the dataset randomly 20 times, and the gallery for testing has 100 identities each time.
We only randomly split the dataset once for training and testing.
In addition, we mainly report CMC accuracy and did not consider mAP because of the different gallery size.

The results are shown in Table. \ref{table:Market-1501} $\sim$ \ref{table:CUHK-SYSU}.
For the SCPNet, we choose the results of the experiment with $\lambda = 10$ and $R=4$.
Overall, our proposed SCPNet achieves the best performance on Market-1501, CUHK-SYSU and CUHK03, and lags behind the best results with a slight gap on DukeMTMC-ReID.
It is worth mentioning that DSR also reported its results on Market1501, which our SCPNet outperforms by $10.5\%$ rank-1 accuracy and $17.5\%$ mAP.
In conclusion, the SCPNet can perform well on the holistic person ReID task.

\subsubsection{Results on Partial ReID.}

We compare the proposed SCPNet to the state-of-the-art methods, including AMC, SWM, AMC+SWM, DSR and Resizing model, on the Partial REID and Partial-iLIDS datasets.
There are $p=60$ and $p=119$ individuals in each of the test sets for Partial REID and Partial-iLIDS datasets respectively.
The state-of-the-art results are taken from \cite{he2018deep} and more details can be found in it.
For the SCPNet, $\lambda$ is set to $10$ and $R$ is $4$.
Note that the state-of-the-art results are achieved by supervised learning, while our SCPNet has not been trained on the Partial REID and Partial-iLIDS datasets.
As Table. \ref{table:Partial REID} shown, our SCPNet finally achieves $68.3 \%$ and $84.9\%$ rank-1 accuracy on Partial REID and Partial-iLIDS respectively.
These unsupervised cross-domain results beat existing supervised learning methods by a large margin ($25.3 \%$ and $30.3\%$ rank-1 accuracy on Partial REID and Partial-iLIDS respectively).
However, the \emph{Baseline} ($R=4, \lambda=0$) also performs better than existing methods.
This surprising result gives us an inspiration.
We can use the holistic person images to train a partial ReID model, because holistic person images are more easily collected than partial person images.
And with our proposed spatial-channel parallelism, the ReID model can be more suitable for partial person images without extra computational cost in the inference stage.

\begin{threeparttable}[tb]
    %\begin{table}[htb]
    %\begin{center}
    \caption{Experimental results on Partial REID Datasets with single query ($N=1$)}
    \label{table:Partial REID}
    \renewcommand{\arraystretch}{1.2}
    \begin{tabular}{c|c|ccc|ccc}
        \hline
        &	& \multicolumn{3}{c|}{Partial REID, $p=60$} & \multicolumn{3}{c}{Partial-iLIDS, $p=119$}\\
        Method 	& Type	& r = 1 & r = 5 & r = 10 	& r = 1 & r = 5 & r = 10\\
        \hline
        \hline
        Resizing model	 &supervised & 19.3&40.0&51.3		&21.9&43.7	&55.5\\
        SWM \cite{Zheng_2015_ICCV}&supervised& 24.4 & 52.3 & 61.3	& 33.6 & 53.8 &63.3 \\
        AMC \cite{Zheng_2015_ICCV}&supervised& 33.3 & 52.0 & 62.0	& 46.8 & 69.6 &81.9 \\
        AMC+SWM \cite{Zheng_2015_ICCV}&supervised& 36.0 & 60.0 & 70.7	& 49.6 & 72.7 &84.7 \\
        DSR (Multi-scale)\cite{he2018deep}&supervised&  43.0 & 75.0 & 76.7	& 54.6 & 73.1 &85.7 \\
        \hline
        Baseline &unsupervised& 60.0 & 78.3 & 83.7 & 78.2 & 89.1 & 92.4\\
        SCPNet-a \tnote{\textdagger} &unsupervised & 56.3 & 73.3 & 80.5 & 69.8 & 89.9 & \textbf{95.0} \\
        SCPNet-a &unsupervised & \textbf{68.3} & \textbf{80.7} & \textbf{88.3} & \textbf{84.9} & \textbf{92.4} & 94.1 \\
        \hline
    \end{tabular}
    %\end{center}
    \begin{tablenotes}
        \item[\textdagger] After the initial version of this paper, \cite{he2018deep} releases a new evaluation protocol, and we also provide scores under the new protocol here.
    \end{tablenotes}
    %\end{table}
\end{threeparttable}

%===========================================================
\section{Conclusions}
In this paper, we propose a two-branch deep CNN network called Spatial-Channel Parallelism Network (SCPNet) with a local feature branch and a global feature branch. The local feature branch output local features without a global view, while the global feature branch output global features with heavily coupling and redundancy. With spatial-channel parallelism, the channel feature branch can learn local spatial feature through the guidance of local feature branch, and the re-learned local features are more discriminative with a global view, achieving a balance between global and local representation. In this way, the network can also extract local features automatically from the whole image which makes it also more suitable for partial ReID. A single SCPNet model is trained end-to-end on four holistic ReID datasets and achieves the state-of-the-art results on all four datasets. Furthermore, the trained model also outperforms the state-of-the-art results on two partial ReID datasets by a large margin without training.
\clearpage


\bibliographystyle{splncs04}
\bibliography{egbib}

%Author, F.: Article title. Journal \textbf{2}(5), 99--110 (2016)
%Author, F., Author, S.: Title of a proceedings paper. In: Editor,
%F., Editor, S. (eds.) CONFERENCE 2016, LNCS, vol. 9999, pp. 1--13.
%Springer, Heidelberg (2016). \doi{10.10007/1234567890}
%Author, F., Author, S., Author, T.: Book title. 2nd edn. Publisher,
%Author, A.-B.: Contribution title. In: 9th International Proceedings
%on Proceedings, pp. 1--2. Publisher, Location (2010)
%LNCS Homepage, \url{http://www.springer.com/lncs}. Last accessed 4
%\end{thebibliography}

\end{document}